\documentclass[11pt]{article}

\usepackage[preprint]{acl}

\usepackage{times}
\usepackage{latexsym}

\usepackage[T1]{fontenc}

\usepackage[utf8]{inputenc}
\usepackage{booktabs}

\usepackage{microtype}

\usepackage{inconsolata}

\usepackage{graphicx}
\usepackage{amsmath}
\usepackage{url}

\title{LLM Consumer Behavior Theory: Foundations of a Novel Research Field}

  \author{Manon Reusens$^{1}$\thanks{Equal contribution}, Sofie Goethals$^{1}$\footnotemark[1], David Martens$^{1}$ \\
         $^1$Department of Engineering Management, University of Antwerp \\ %
         \small{\{manon.reusens, sofie.goethals, david.martens\}@uantwerpen.be }}

\begin{document}
\maketitle
\begin{abstract}
Large language models (LLMs) are increasingly deployed as autonomous agents that make consumption decisions on behalf of users. This shift raises fundamental questions for consumer theory, which has traditionally modeled humans as the primary decision-makers. In this paper, we introduce \textbf{LLM Consumer Behavior Theory}, a new field of study concerned with analyzing consumer behavior in agentic markets.
Drawing on classical and behavioral economics alongside recent advances in Natural Language Processing, we formalize how human preferences are reflected and acted upon by LLM-based agents, and how agent-level decisions aggregate into market demand. We unify previously fragmented literature on LLM decision-making, human behavior simulation, and preference elicitation under a common economic lens, highlighting where assumptions, such as rationality and heterogeneity, may fail in agentic markets.
Rather than providing empirical validation, this paper outlines the scope of LLM consumer behavior and identifies open research questions related to alignment, preference representation, and market dynamics. %
\end{abstract}

\section{Introduction}
AI systems have undergone a shift from passive conversational tools to \textit{agentic} AI systems that increasingly make decisions on our behalf~\cite{gaarlandt2025aiagents, Purdy2024AgenticAI,goli2024llmscapturehumanpreferences}. These emerging AI agents can proactively plan, evaluate alternatives, and execute complex tasks across domains~\citep{acharya2025agentic,whiting2024aiagents}. %
 From an economic point of view, these systems are a new type of actors increasingly making purchasing decisions on our behalf~\citep{cherep2025framework}. This is also highlighted by the anticipated growth of agentic commerce: McKinsey Research, for example, estimates that by 2030 agentic commerce could orchestrate \$ 3 to \$ 5 trillion globally~\citep{schumacher2025agentic}. Accordingly, classical economic frameworks that solely model individuals as decision-makers must evolve to account for agentic systems acting on behalf of users.

\begin{figure*}
    \centering
    \includegraphics[width=0.8\linewidth]{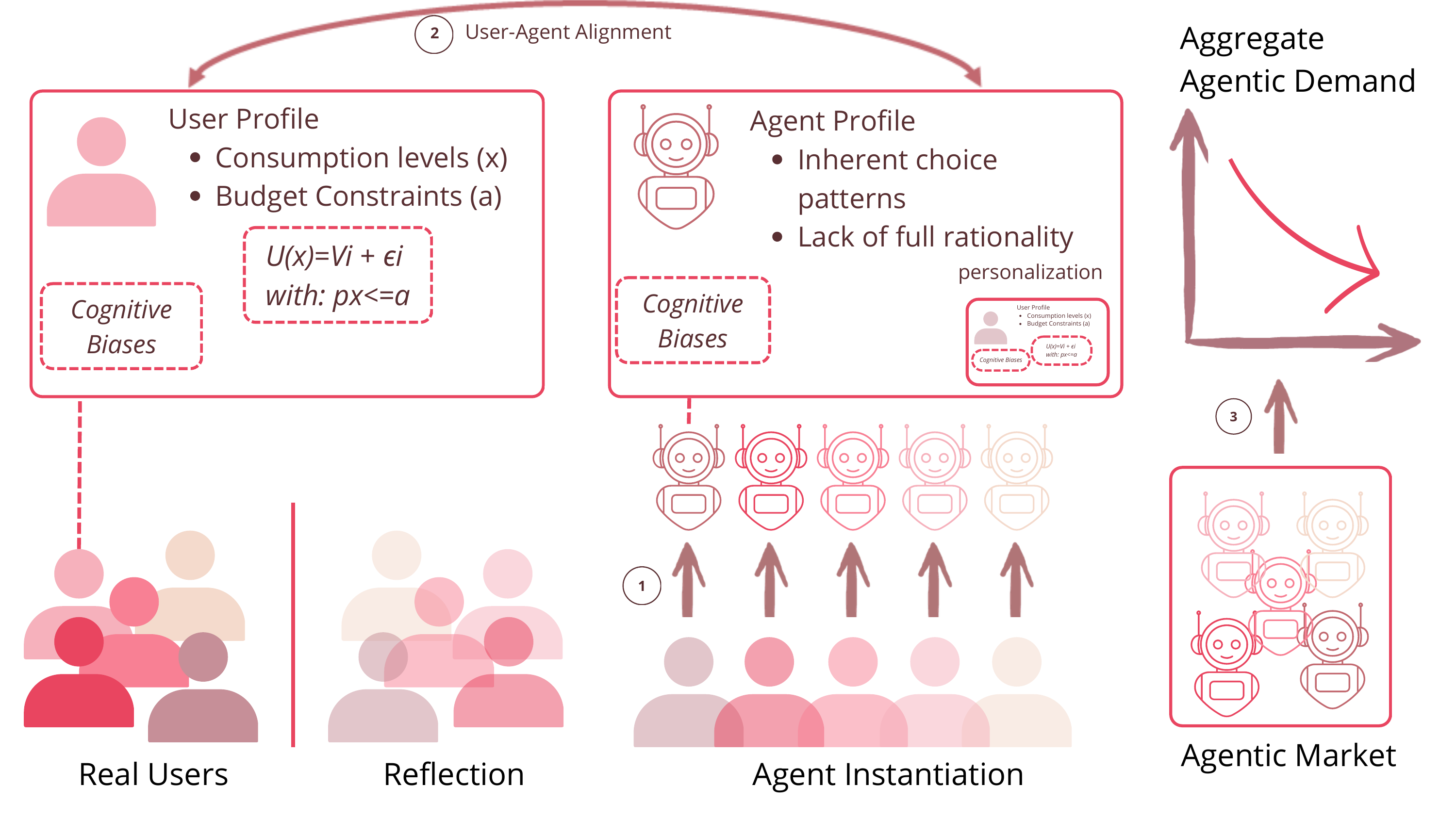}
    \caption{Field of LLM Consumer Behavior Theory. Consumers are modeled via preference structures. These preferences are imperfectly reflected and instantiated in LLM agents. Each agent acts on behalf of a user, producing agent‑level choices that aggregate into market demand}
    \label{fig:agentic_market}
\end{figure*}

This paper introduces \textbf{LLM Consumer Behavior Theory}, a new research field concerned with consumption decisions made by LLM-based agents acting on behalf of human users. By integrating insights from classical consumer behavior theory, behavioral economics, and Natural Language Processing (NLP), we unify previously fragmented literature under a common economic framework, establishing a coherent foundation for studying agentic consumer behavior. Rather than exhaustively characterizing all implications of this emerging domain, we outline the scope and core conceptual structure of LLM consumer behavior, clarify how existing empirical findings fit within it, and identify central open questions that arise when consumer decision-making is delegated to algorithmic agents.

The paper is organized around three core components of LLM Consumer Behavior Theory, summarized in Figure~\ref{fig:agentic_market}.
First, we ground \textbf{(LLM) consumer preferences} in classical utility theory and behavioral economics, analyzing how choices arise from preferences, constraints, and cognitive biases.
Second, we study \textbf{the user–agent alignment}, formalizing how consumer preferences are reflected in LLM-based agents through an agent–principal perspective.
Third, we turn to \textbf{consumer markets}, aggregating agent-level decisions to characterize agentic demand and exploring how heterogeneity and market dynamics might be affected in comparison to human consumer markets, and how alignment and diversification can mitigate these effects.

\section{Consumer preferences and utility}

This section reviews the principal theories used to model \textit{individual} consumers, and discusses related work on how these frameworks have been applied in autonomous decision-making settings.

\subsection{Model of Individual Choice} \
In classical economic theory, consumers are assumed to choose the option that maximizes utility, represented by $U(x)$, where $x$ is a set of consumption levels, subject to a budget constraint $px \leq a$ where $p$,where $p$ is a set of prices and $a$ is the income of the consumer~\cite{mcfadden2001economic}. This framework is commonly referred to as the rational choice model~\cite{alma9993574800201488}.  In practice, however, decision-making may also be influenced by random factors, such as noise or bias. Therefore, random utility theory was introduced to analyze choices that maximize utility, while allowing for random elements~\cite{mcfadden2001economic}. Utility is then decomposed into a systematic part $V$ and an error term or unobservable part $\epsilon$~\cite{MASIERO2015117}:
\begin{equation} 
    U(x)=V+\epsilon
\end{equation}
Different models can approximate this utility function, such as the multinomial logit model or the mixed logit model \cite{MASIERO2015117,mcfadden2001economic,reusens2026largelanguagemodelpay}

Research in NLP has begun to study LLM choice preferences in market-like settings. \citet{cedro2025cash}  study a series of real-world dilemmas, such as whether to accept a discount in exchange for longer waiting times, and map the resulting preferences of several LLMs. They find that smaller and older models exhibit less consistent preferences, whereas larger and more recent models display sharper and more stable patterns. Related studies consider other forms of LLM decision-making, including preferences over time ~\cite{goli2024llmscapturehumanpreferences}, risk ~\cite{jia2024decision}, or walking distance~\cite{FULMAN2025104542}.
From an economic perspective, \citet{reusens2026largelanguagemodelpay} use a discrete choice method to estimate Willingness to Pay for hotel room features.\footnote{The maximum willingness to pay signals that the consumer is indifferent between buying or not buying the product~\cite{alma9993574800201488}}.

\subsection{Behavioral Economics}
Classical economic theory treats the systematic component of human decision-making as rational. Yet even within this component, people are influenced by behavioral biases and do not behave fully rationally. Deviations include loss aversion, anchoring, overconfidence, and framing effects~\citep{tversky1979prospect,kahneman2011thinking}.

\paragraph{LLM Rationality} \citet{jiang-2025-towards} characterizes a rational agent using four main axioms: information grounding, logical consistency, invariance from irrelevant context, and orderability of preferences. Evidence suggests that LLMs do not fully satisfy these criteria, due in part to inconsistent outputs, a bounded knowledge space, and their lack of direct real-world grounding and feedback mechanisms ~\cite{jiang-etal-2025-towards}. At the same time, some evidence of economic rationality is found, although these are highly sensitive to prompt variations~\cite{chen2023emergence}. A growing body of work thus investigates how rationality can be improved through, for example multimodal systems, integrated tool use, or additional alignment~\cite{jiang-2025-towards,liu-etal-2025-evaluating}. However, these approaches do not yet succeed in making LLMs fully rational~\cite{jiang-2025-towards,liu-etal-2025-evaluating}.

\paragraph{LLM cognitive biases} Beyond violations of rationality, research examines whether LLMs exhibit systematic behavioral deviations similar to those observed in humans~\citep{liu-etal-2025-evaluating,jones2022capturing,jia2024decision}. 
These models apply higher discount rates than humans in intertemporal tasks while simultaneously producing more internally consistent choices in gambling-like settings \citep{goli2024llmscapturehumanpreferences,liu2025large}. \citet{jones2022capturing} show that prompt designs inspired by human cognitive biases can reliably induce systematic errors. Additionally, both risk and loss aversion are present within LLMs, though the degree is model-dependent~\cite{jia2024decision} and LLMs are more susceptible to nudges than humans~\cite{cherep2025llm}.
In general, LLMs exhibit a range of behaviors between these human biases and more economically rational decisions~\cite{chen2023emergence,raman2024steer,ross2024llm, bini2025behavioral, lou2026anchoring}.

\section{User-Agent Alignment}\label{subsec:user_agent_alignment}
In agentic markets, LLMs make decisions on behalf of users, requiring alignment with user preferences. Assuming one agent per user, this setting maps onto the classical principal-agent problem, where decision authority is delegated under imperfect information and potentially divergent objectives~\cite{GrossmanSanfordJ.1983AAot}. Users delegate decisions to LLMs expecting that these systems act in their best interest. Yet LLMs are shaped by the design choices of their developers, which can lead to misalignment between model behavior and the behavior that a user might seek, as shown by \citet{Fan_Chen_Jin_He_2024} and \citet{reusens-etal-2025-economists}. Additionally, LLMs do not directly observe true user preferences, but rely on proxy representations inferred from prompts, context history, past behavior, and available demographic information (Figure~\ref{fig:agentic_market}).

\paragraph{Persona-based alignment} Research studies the use of LLMs to simulate human behavior~\cite{wang-etal-2025-know,yoon-etal-2024-evaluating,kim2026llmlearnpreferenceschoice,horton2023large}. A popular approach for improving human alignment is \textit{persona} assignment: instructing an LLM to ``\textit{act as}" a user with specific characteristics or preferences~\cite{wang-etal-2025-know}. Personas can also be based on behavioral rather than demographic attributes%
~\citep{goethals2025words}. While persona assignment can improve alignment in simple tasks~\cite{liu-etal-2025-evaluating}, it also has important limitations: models may not follow assigned personas consistently~\cite{horton2023large, reusens-etal-2025-economists}, especially when asked to follow uncommon preferences~\cite{Fan_Chen_Jin_He_2024}, and may behave stereotypically~\cite{wang2025large,reusens-etal-2025-economists}. Lastly, careful persona design is required as it could lead to extreme model behavior~\cite{reusens2026largelanguagemodelpay}.  %

\paragraph{Alignment through few-shot learning and finetuning} A complementary line of work studies whether LLM-based agents can learn individual preferences from limited interaction data. Several papers use in-context-learning from past choices i.e. \cite{kim2026llmlearnpreferenceschoice,singh2025fspo,reusens2026largelanguagemodelpay}. Other papers finetune a user-specific personalized model i.e.~\cite{thonet-etal-2025-fast}. However, full fine-tuning per user can be costly and hard to scale. Therefore, more efficient methods are also proposed, such as LoRe, that models individual preferences as weighted combinations of base reward functions~\cite{boselore}. %

\section{Consumer Markets}\label{market}
If everyone uses an agent as their proxy, a fully agentic consumer market would emerge, as shown in Figure~\ref{fig:agentic_market}. In microeconomics, consumer's Willingness to Pay values constitute the demand governed by the \emph{Law of Demand}: as the quantity sold increases, the price must fall to attract additional buyers~\cite{marshall1961principles}. In agentic markets, aggregating individual agentic willingness to pay values across this population thus yields the agentic market demand curve.

\subsection{The Heterogeneous User Demand}

Individual human choices are driven by utility, which depends on factors such as preferences and income~\cite{marshall1961principles} but also on unobserved dimensions such as tastes and perceptions of the world~\cite{mcfadden2001economic}. As a result, traditional consumer markets are characterized by substantial heterogeneity in preferences reflecting differences in income, taste, and lived experiences. This heterogeneity is a central feature of economic models, as variation across individuals shapes both market outcomes and welfare implications~\citep{mcfadden1972conditional, berry1993automobile}.

\subsection{The Homogenous Agentic Demand }
\label{sec:homogenization}
Heterogeneity may be reduced in agentic markets, since LLMs tend to reproduce average patterns from their training data rather than diverse human preferences~\citep{bender2021dangers, matz2025basic}. 
This effect may be reinforced by the concentrated LLM ecosystem, in which state-of-the-art models rely on largely overlapping corpora, similar architectures, and related alignment procedures~\citep{bommasani2021opportunities, intelmarketresearch2024llm}.
Although homogenization has been documented in other downstream applications~\citep{weidinger2021ethical, shumailov2024ai, goethals2025one}, its implications for consumer markets and aggregate demand remain underexplored.
Early evidence suggests that agentic demand may be more homogeneous than human demand: LLMs show more uniform out-of-the-box choice patterns~\cite{chen2023emergence}, AI agents can concentrate demand on a few ``modal'' products~\cite{allouah2025your}, and LLM behavior appears less heterogeneous than human behavior~\cite{del2025can}. These preferences may also be unstable, with model updates substantially shifting market shares~\cite{allouah2025your}.

\subsection{Diversifying the Agentic Demand}
Agentic heterogeneity can still arise from both model-side and user-side factors. On the model side, architectural differences across foundation models can lead to different decisions or price boundaries~\citep{reusens2026largelanguagemodelpay,cedro2025cash}, and higher temperature settings can also induce response diversity~\cite{chen2023emergence}. On the user side, aligning agents to specific user profiles can further diversify behavior, as discussed in Section~\ref{subsec:user_agent_alignment}. For example, LLMs can behave like heterogeneous households when assigned different profiles, occupations, and income levels, generating consumption patterns consistent with macroeconomic regularities \cite{li-etal-2024-econagent}.

\section{Discussion}
A central implication of LLM Consumer Behavior Theory is that consumer analysis can no longer focus solely on human users. When decisions are delegated to LLM-based agents, market outcomes reflect a broader pipeline involving preference reflection, initial agentic preferences, and user–agent alignment (Figure~\ref{fig:agentic_market}). Understanding behavior along this continuum is therefore critical for analyzing emerging consumer markets, motivating several directions for future research.

\paragraph{LLM choice patterns}
A first key direction concerns LLM choice patterns. What factors influence the choices these models make? How robust and consistent are these choices across models, settings and prompts? To what extent can they be influenced or steered, either through prompting strategies or through modifications to the model itself?%

\paragraph{Personalized agents}
A second direction concerns aligning agents with unique user preferences. Which data sources and alignment mechanisms best capture true user preferences?
How do reflection errors or biases propagate from agent decisions to market-level distortions? 
Can shared agents effectively serve heterogeneous users without disadvantaging minority groups, or does meaningful alignment require individualized models?

\paragraph{Hybrid Markets}
Rather than an abrupt shift, consumer markets are likely to evolve gradually from human-driven to increasingly agentic systems, creating hybrid market structures. Open questions include the pace of this transition, the dynamics of hybrid markets, and whether shared model weights could induce preference homogenization without collusion. What risks arise when large populations of agents update simultaneously, and how will human consumers adapt to the growing concentration of agentic decision-making? How are prices and equilibria affected in hybrid markets?

\section{Conclusion}
This paper introduces \textbf{LLM Consumer Behavior Theory}, which studies consumption decisions made by LLM-based agents acting on behalf of human users. As consumer decision-making becomes increasingly agentic, market behavior can no longer be understood solely through human preferences. By grounding this in classical and behavioral consumer theory and linking it to recent advances in NLP, we clarify how empirical findings on LLM behavior map to established economic concepts and why agentic consumption raises new challenges of alignment and heterogeneity. We delineate the scope of this emerging field and identify foundational questions for future research. %

\section*{Limitations}
As noted, the primary aim of this paper is to introduce LLM Consumer Behavior Theory as a novel field of study, rather than to provide empirical validation, which we leave to future work. Our analysis is further limited to the demand side of agentic markets; extending these ideas to agentic supply, and to settings in which both supply and demand are partially agentic, raises additional questions for future research.  Finally, we do not attempt an exhaustive review of all related literature. Instead, we synthesize insights across economics and NLP to offer a structured perspective on agentic consumer behavior.

\section*{Ethical Considerations}
Analyzing agentic consumer markets also raises several ethical considerations. A central issue concerns \textit{accountability} when LLMs make decisions on behalf of users. Who is to blame for a poor decision? Currently, there is no global consensus for this question~\citep{schumacher2025agentic}.
Individuals should provide informed consent and understand to what extent they remain responsible for agentic choices. This concern extends beyond fully autonomous settings to cases where users explicitly approve recommendations, as approving a suggestion generated by an LLM involves more cognitive engagement and deliberation than making the decision independently. %
Additional ethical risks arise when agents are misaligned with user preferences and exhibit stereotypical or biased behavior. Finally, the complexity of LLMs limits transparency into their underlying decision-making processes. Addressing these challenges represents an important direction for future research.

\section*{Acknowledgments}
We would like to thank the Antwerp Center on Responsible AI (ACRAI) for their support.

\bibliography{anthology,custom}

\end{document}